%% file: main.tex
\documentclass[runningheads]{llncs}

\usepackage[utf8]{inputenc}
\usepackage{amsfonts}
\usepackage{amsmath}
\usepackage{tikz}
\usepackage{listings}

\input{macros}
\usepackage{multirow}

\title{Certifying Decision Trees Against Evasion Attacks by Program Analysis\thanks{Accepted to ESORICS 2020}}

\author{Stefano Calzavara \and Pietro Ferrara \and Claudio Lucchese}
\institute{Università Ca' Foscari Venezia}

\begin{document}

\maketitle

\begin{abstract}
Machine learning has proved invaluable for a range of different tasks, yet it also proved vulnerable to evasion attacks, i.e., maliciously crafted perturbations of input data designed to force mispredictions. In this paper we propose a novel technique to verify the security of decision tree models against evasion attacks with respect to an expressive threat model, where the attacker can be represented by an arbitrary imperative program. Our approach exploits the interpretability property of decision trees to transform them into imperative programs, which are amenable for traditional program analysis techniques. By leveraging the abstract interpretation framework, we are able to soundly verify the security guarantees of decision tree models trained over publicly available datasets. Our experiments show that our technique is both precise and efficient, yielding only a minimal number of false positives and scaling up to cases which are intractable for a competitor approach.

\end{abstract}

\input{intro}
\input{background}
\input{proposal}
\input{implementation}
\input{experiments}
\input{related}
\input{conclusion}

\bibliographystyle{splncs}
\bibliography{biblio}

\end{document}

%% file: macros.tex
\newcommand{\R}{\mathbb{R}}
\newcommand{\feats}{\mathcal{X}}
\newcommand{\labels}{\mathcal{Y}}
\newcommand{\dataset}{\mathcal{D}}
\newcommand{\hyps}{\mathcal{H}}
\DeclareMathOperator*{\argmin}{argmin}
\newcommand{\Loss}{\mathcal{L}}
\newcommand{\dtrain}{\dataset_{train}}
\newcommand{\dtest}{\dataset_{test}}
\renewcommand{\vec}[1]{\mathbf{#1}}
\newcommand{\atkloss}{\Loss^A}
\newcommand{\leaf}[1]{\lambda(#1)}
\newcommand{\node}[1]{\sigma(#1)}
\newcommand{\rewrite}[4]{#1 \xrightarrow{#2}_{#3} #4}

%% file: intro.tex
\section{Introduction}
Machine learning (ML) learns predictive models from data and has proved invaluable for a range of different tasks, yet it also proved vulnerable to \emph{evasion attacks}, i.e., maliciously crafted perturbations of input data designed to force mispredictions~\cite{SzegedyZSBEGF13}. To exemplify, let us assume a credit company decides to use a ML model to automatically assess whether customers qualify for a loan or not. A malicious customer who somehow realises or guesses that the model privileges unmarried people over married people could cheat about her marital status to improperly qualify for a loan.

The research community recently put a lot of effort in the investigation of \emph{adversarial} ML, e.g., techniques to train models which are resilient to attacks or assess the security properties of models. In the present paper we are interested in the \emph{security certification} of a popular class of models called \emph{decision trees}, i.e., we investigate formally sound techniques to quantify the resilience of such models against evasion attacks. Specifically, we propose the first \emph{provably sound} certification technique for decision trees with respect to an expressive threat model, where the attacker can be represented by an arbitrary imperative program. Verifying ML techniques with respect to highly expressive threat models is nowadays one of the most compelling research directions of adversarial ML~\cite{MoreRobustVerification,semanticattacker}. This is an important step forward over previous work, which either proposed empirical techniques without formal guarantees or only focused on artificial attackers expressed as mathematical distances (see Section~\ref{sec:related} for full details).

Our approach exploits the \emph{interpretability} property of decision trees, i.e., their amenability to be easily understood by human experts, which makes their translation into imperative programs a straightforward task. Once a decision tree is translated into an imperative program, it is possible to leverage state-of-the-art program analysis techniques to certify its resilience to evasion attacks. In particular we leverage the \emph{abstract interpretation} framework~\cite{CC77,CC79} to automatically extract a sound abstraction of the behaviour of the decision tree under attack. This allows us to efficiently compute an over-approximated, yet precise, estimate of the resilience of the decision tree against evasion attacks.

\subsubsection{Contributions.}
We specifically contribute as follows:
\begin{enumerate}
    \item We propose a general technique to certify the security guarantees of decision trees against evasion attacks attempted by an attacker expressed as an arbitrary imperative program. We exemplify the technique at work on an expressive threat model based on rewriting rules (Section~\ref{sec:proposal}).
    \item We implement our technique into a new tool called TreeCert. Given a decision tree, an attacker and a test set of instances used to estimate prediction errors, TreeCert outputs an over-approximation of the error rate that the attacker can force on the decision tree. TreeCert implements a \emph{context-insensitive} analysis computing a single over-approximation of the attacker's behavior and reuses it in the analysis of all the test instances, thus boosting efficiency without missing attacks (Section~\ref{sec:implementation}).
    \item We experimentally prove the effectiveness of TreeCert against publicly available datasets. Our results show that TreeCert is extremely precise, since it can compute tight over-approximations of the actual error rate under attack, with a difference of at most 0.02 over it on cases which are small enough to be analyzed without approximated techniques. Moreover, TreeCert is much faster than a competitor approach~\cite{CalzavaraLT19} and scales to intractable cases, avoiding the exponential blow-up of non-approximated techniques (Section~\ref{sec:experiments}).
\end{enumerate}

%% file: background.tex
\section{Background}

\subsection{Security of Supervised Learning}
In this paper, we deal with the security of \emph{supervised learning}, i.e., the task of learning a classifier from a set of labeled data. Formally, let $\feats \subseteq \R^d$ be a $d$-dimensional space of real-valued features and $\labels$ be a finite set of class labels; a \emph{classifier} is a function $f: \feats \rightarrow \labels$ which assigns a class label to each element of the vector space (also called \emph{instance}). The correct label assignment for each instance is modeled by an unknown function $g: \feats \rightarrow \labels$, called \emph{target} function. 

Given a \emph{training set} of labeled data $\dtrain = \{(\vec{x}_1,g(\vec{x}_1)),\ldots,(\vec{x}_n,g(\vec{x}_n))\}$ and a \emph{hypothesis space} $\hyps$, the goal of supervised learning is finding the classifier $\hat{h} \in \hyps$ which best approximates the target function $g$. Specifically, we let $\hat{h} = \argmin_{h \in \hyps} \Loss(h, \dtrain)$, where $\Loss$ is a \emph{loss} function which estimates the cost of the prediction errors made by $h$ on $\dtrain$. Once $\hat{h}$ is found, its performance is assessed by computing $\Loss(\hat{h},\dtest)$, where $\dtest$ is a \emph{test set} of labeled, held-out data drawn from the same distribution of $\dtrain$.

Within the context of security certification, one should measure the accuracy of $\hat{h}$ by taking into account all the actions that an attacker could take to fool the classifier into mispredicting, a so-called \emph{evasion attack}~\cite{BiggioCMNSLGR13,BiggioR18}. To provide a more accurate evaluation of the performance of the classifier under attack, the loss $\Loss$ can thus be replaced by the \emph{loss under attack} $\atkloss$~\cite{MadryMSTV18}. Formally, the attacker can be modeled as a function $A: \feats \rightarrow 2^\feats$ mapping each instance into a set of \emph{perturbed} instances which might fool the classifier. The test set $\dtest$ can thus be corrupted into any dataset obtained by replacing each $(\vec{x}_i,y_i) \in \dtest$ with any $(\vec{x}_i',y_i)$ such that $\vec{x}_i' \in A(\vec{x}_i)$; we let $A(\dtest)$ stand for the set of all such datasets. The loss under attack $\atkloss$ is thus defined by making the pessimistic assumption that the attacker is able to craft the most damaging perturbations, as follows:
\begin{equation}
\atkloss(\hat{h},\dtest) = \max_{\dataset' \in A(\dtest)} \Loss(\hat{h}, \dataset').
\end{equation}

Unfortunately, computing $\atkloss$ by enumerating $A(\dtest)$ is intractable, given the huge number of perturbations available to the attacker: for example, if the attacker can flip $K$ binary features, then each instance can be perturbed in $2^K$ different ways, leading to $2^K \cdot |\dtest|$ possible attacks.

\subsection{Decision Trees}
A powerful set of hypotheses $\hyps$ is the set of the \emph{decision trees}~\cite{BreimanFOS84}. We focus on traditional binary decision trees, whose internal nodes perform thresholding over feature values. Such trees can be inductively defined as follows: a decision tree $t$ is either a leaf $\leaf{\hat{y}}$ for some label $\hat{y} \in \labels$ or a non-leaf node $\node{f,v,t_l,t_r}$, where $f \in [1,d]$ identifies a feature, $v \in \R$ is the threshold for the feature $f$ and $t_l,t_r$ are decision trees. At test time, an instance $\vec{x} = (x_1,\ldots,x_d)$ traverses the tree $t$ until it reaches a leaf $\leaf{\hat{y}}$, which returns the \textit{prediction} $\hat{y}$, denoted by $t(\vec{x}) =  \hat{y}$. Specifically, for each traversed tree node $\node{f,v,t_l,t_r}$, $\vec{x}$ falls into the left tree $t_l$ if $x_f \leq v$, and into the right tree $t_r$ otherwise. 

\begin{figure}[t]
\centering
\begin{tikzpicture}
\node[circle,draw,minimum size=4em](z){$x_2 \leq 10$}
  child{node[circle,draw,minimum size=4em]{$x_1 \leq 5$} child{node[circle,draw,minimum size=2.5em] {$+1$}} child{node[circle,draw,minimum size=2.5em] {$-1$}}} 
  child{node[circle,draw,,minimum size=2.5em]{$+1$}  };
\end{tikzpicture}
\caption{Example of decision tree}
\label{fig:dt}
\end{figure}
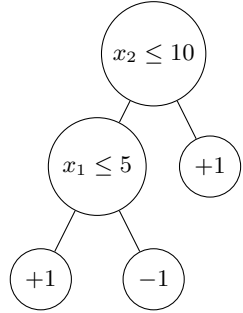

Figure~\ref{fig:dt} represents an example decision tree, which assigns the instance (6,8) with label $-1$ to its correct class. In fact, (i) the first node checks whether the second feature (whose value is 8) is lesser than or equal to 10 and then takes the left sub-tree, and (ii) the second node checks whether the first feature (whose value is 6) is lesser than or equal to 5 and then takes the right leaf, classifying the instance with label $-1$. However, note that an attacker who was able to corrupt (6,8) into (5,8) could force the decision tree into changing its output, leading to the prediction of the wrong class $+1$. 

\subsection{Abstract Interpretation}
In the abstract interpretation framework, the concrete behavior of a program is approximated through \emph{abstract values} of a given \emph{abstract domain} with a lattice structure, rather than concrete values. For example, the Sign domain abstracts numbers with their sign, as formalized by the following \emph{abstraction} and \emph{concretization} functions ($\alpha$ and $\gamma$ respectively):
\[
\alpha(V) = 
\begin{cases}
\bot & \text{if } V = \emptyset \\
+ & \text{if } \forall v \in V :  v > 0 \\
0 & \text{if } \forall v \in V : v = 0 \\
- & \text{if } \forall v \in V : v < 0 \\
\top & \text{otherwise}\\
\end{cases} \quad\quad
\gamma(a) = 
\begin{cases}
\R & \text{if } a = \top \\
\{n \in \R ~|~ n > 0\} & \text{if } a = + \\
\{0\} & \text{if } a = 0 \\
\{n \in \R ~|~ n < 0\} & \text{if } a = - \\
\emptyset & \text{if } a = \bot
\end{cases}
\]

Notice that for all sets of concrete values $V \subseteq \R$ we have $V \subseteq \gamma(\alpha(V))$, i.e., the abstraction function provides an over-approximation of the concrete values. Operations over concrete values like the sum operation $+$ are over-approximated by abstract counterparts $\oplus$ over the abstract domain, which define the \emph{abstract semantics}. For example, the sum of two positive numbers is certainly positive, while the sum of a positive number and a negative number can be positive, negative or 0; this lack of information is modeled by $\top$. Hence, $\oplus$ is defined such that $+\, \oplus\, + = +$ and $+\, \oplus\, - = \top$. A sound definition of $\oplus$, here omitted, must ensure that $\forall V_1 ,V_2 \subseteq \R: \{ v_1 + v_2 ~|~ v_1 \in V_1 \land v_2 \in V_2 \} \subseteq \gamma(\alpha(V_1) \oplus \alpha(V_2))$, i.e., abstract operations must over-approximate operations over concrete values. By simulating the program over the abstract domain, abstract interpretation ensures a fast convergence to an over-approximation of all the reachable program states. In particular, the analysis consists in computing the fixpoint of the abstract semantics over the abstract domain, making use of a \emph{widening} operator -- usually if the upper bound operator does not converge in a given threshold~\cite{CC77,CC79}.

Thanks to its modular approach, abstract interpretation allows one to define multiple abstractions of the same concrete domain. Therefore, several abstract domains approximating numerical values have been proposed in the literature. For instance Octagons~\cite{Octagons} and Polyhedra~\cite{Polyhedra} track different types of (linear) relations among numerical variables, and have been applied to different contexts. Apron~\cite{Apron} is a library of numerical abstract domains comprising the main domains leveraged in this work.

%% file: proposal.tex
\section{Security Verification of Decision Trees}
\label{sec:proposal}

\subsection{Threat Model}
\label{sec:threat}
Our approach is general enough to be applied to attackers represented as arbitrary imperative programs. To exemplify, we show how it can be applied to an expressive threat model based on \emph{rewriting rules}~\cite{CalzavaraLT19}. This relatively new threat model goes beyond traditional distance-based models, which are plausible for perceptual tasks like image recognition, but are inappropriate for non-perceptual tasks (e.g., loan assignment) where mathematical distances do not capture useful semantic properties of the domain of interest. 

We model the attacker $A$ as a pair $(R, K)$, where $R$ is a set of \emph{rewriting rules}, defining how instances can be corrupted, and $K \in \R^+$ is a \emph{budget}, limiting the amount of alteration the attacker can apply to each instance. Each rule $r \in R$ has form:
\begin{equation*}
\rewrite{[a,b]}{f}{k}{[\delta_l,\delta_u]},
\end{equation*}
where: $(i)$ $[a,b]$ and $[\delta_l,\delta_u]$ are intervals on $\R \cup \{-\infty,+\infty\}$, with the former defining the \emph{precondition} for the application of the rule and the latter defining the \emph{magnitude} of the perturbation enabled by the rule; $(ii)$ $f \in [1,d]$ is the index of the feature to perturb; and $(iii)$ $k \in \R^+$ is the \emph{cost} of the rule. The semantics of the rewriting rule can be explained as follows: if an instance $\vec{x} = (x_1,\ldots,x_d)$ satisfies the condition $x_f \in [a,b]$, then the attacker can corrupt it by adding any $v \in [\delta_l,\delta_u]$ to $x_f$ and spending $k$ from the available budget. The attacker can corrupt each instance by using as many rewriting rules as desired in any order, possibly multiple times, up to budget exhaustion. 

According to this attacker model, we can define $A(\vec{x})$, the set of the attacks against the instance $\vec{x}$, as follows.

\begin{definition}[Attacks]
Given an instance $\vec{x}$ and an attacker $A = (R,K)$, we let $A(\vec{x})$ be the set of the \emph{attacks} that can be obtained from $\vec{x}$, i.e., the set of the instances $\vec{x}'$ such that there exists a sequence of rewriting rules $r_1,\ldots,r_n \in R$ and a sequence of instances $\vec{x}_0,\ldots,\vec{x}_n$ where:
\begin{enumerate} 
\item $\vec{x}_0 = \vec{x}$ and $\vec{x}_n = \vec{x}'$;
\item for all $i \in [1,n]$, the instance $\vec{x}_{i-1}$ can be corrupted into the instance $\vec{x}_i$ by using the rewriting rule $r_i$, as described above;
\item the sum of the costs of $r_1,\ldots,r_n$ is not greater than $K$.
\end{enumerate}
Notice that $\vec{x} \in A(\vec{x})$ for any $A$ by picking an empty sequence of rewriting rules.
\end{definition}

\begin{example}
\label{ex:example}
Consider the attacker $A = (\{r_1,r_2\},10)$, where:
\begin{itemize}
    \item $r_1 = \rewrite{[0,10]}{1}{5}{[-1,0]}$ allows the attacker to corrupt the first feature by adding any value in $[-1,0]$, provided that the feature value is in $[0,10]$ and the available budget is at least 5;
    \item $r_2 = \rewrite{[5,10]}{2}{4}{[0,1]}$ allows the attacker to corrupt the second feature by adding any value in $[0,1]$, provided that the feature value is in $[5,10]$ and the available budget is at least 4.
\end{itemize}

The attacker $A$ can force the decision tree in Figure~\ref{fig:dt} to change its original prediction ($-1$) on the instance $(6,8)$. In particular, we can show that $(5,8)$ is a possible attack against $(6,8)$, since $A$ can apply $r_1$ once by spending 5 from the budget, and $(5,8)$ is classified as $+1$ by the decision tree.
\end{example}

\subsection{Conversion to Imperative Program}
\label{subsect:convertiontoimperativeprogram}
Our analysis technique exploits the \emph{interpretability} property of decision trees, i.e., their amenability to be easily understood by human experts. In particular, it is straightforward to convert any decision tree into an equivalent, loop-free imperative program. To exemplify, Figure~\ref{fig:imp} shows the translation of the decision tree in Figure~\ref{fig:dt} into an equivalent function.

\begin{figure}[t]
\begin{lstlisting}
int predict (float[] x) {
    if (x[2] <= 10) {
        if (x[1] <= 5)
            return +1;
        else
            return -1;
    }
    else
        return +1;
}
\end{lstlisting}
\caption{Translation of the decision tree in Figure~\ref{fig:dt} into an imperative program}
\label{fig:imp}
\end{figure}

We can then model the attacker as an imperative program which has access to the function representing the decision tree to analyse. In particular, we observe that the attacker $A = (R,K)$ can be represented by means of a \emph{non-deterministic} program which behaves as follows:
\begin{enumerate}
    \item Select a random rewriting rule $r \in R$.
    \item Let $\rewrite{[a,b]}{f}{k}{[\delta_l,\delta_u]}$ be the selected rule $r$ and let $\vec{x} = (x_1,\ldots,x_d)$ be the instance to perturb. If $x_f \in [a,b]$ and the available budget is at least $k$, then select a random $\delta \in [\delta_l,\delta_u]$, replace $x_f$ with $x_f + \delta$ and subtract $k$ from the available budget.
    \item Non-deterministically go to step 1 or terminate the process. This stop condition allows the attacker to spare part of the budget, which is needed to enforce termination when the entire budget cannot be spent (or does not need to).
\end{enumerate}

This encoding is exemplified in Figure~\ref{fig:imp-att}, where lines 1-27 show how the attacker of Example~\ref{ex:example} can be modeled as an imperative program, using standard functions for random number generation. Once the attacker has been modeled, we can finally encode the behavior of the decision tree under attack: this is shown in lines 29-32, where we let the attacker corrupt the input instance before it is fed to the decision tree for prediction.

\begin{figure}[t]
\lstset{morekeywords={random_int, random_float}}
\begin{lstlisting}
float[] attack (float[] x) {
    float K = 10;
    boolean done = false;
    while (!done) {
        int rule = random_int(1,3);
        switch (rule) {
            case 1:
                if (x[1] >= 0 && x[1] <= 10 && K >= 5) {
                    float delta = random_float(-1,0);
                    x[1] = x[1] + delta;
                    K = K - 5;
                }
                break;
            case 2:
                if (x[2] >= 5 && x[2] <= 10 && K >= 4) {
                    float delta = random_float(0,1);
                    x[2] = x[2] + delta;
                    K = K - 4;
                }
                break;
            case 3:
                // this models non-deterministic termination
                done = true;
        }
    }
    return x;
}

int predict_under_attack (float[] x) {
    float[] x' = attack(x);
    return predict(x');
}
\end{lstlisting}
\caption{Encoding predictions under attack into an imperative program}
\label{fig:imp-att}
\end{figure}

\subsection{Proving Security by Program Analysis}
\label{sec:approx}
Given a decision tree $t$, an attacker $A$ and a test set $\dtest$, we can compute an over-approximation of $\atkloss(t,\dtest)$ as follows. 

We first translate the decision tree $t$ together with the attacker $A$ into an imperative program $P$ modeling the decision tree under attack, as discussed in Section~\ref{subsect:convertiontoimperativeprogram}. For each instance $(\vec{x}_i,y_i) \in \dtest$, we build an abstract state $\alpha(\{\vec{x}_i\})$ representing $\vec{x}_i$ in the chosen abstract domain and we analyze $P$ with such entry state. Then, the output of the analysis might be either of the following:
\begin{enumerate}
    \item only leaves of the decision tree with the correct class label $y_i$ are reachable. This means that, for all possible attacks against $\vec{x}_i$, the decision tree always classifies  the instance correctly;
    \item leaves with the wrong label are reachable as well. If $t$ correctly classifies the instance in the unattacked setting, this might happen either because there is indeed an attack leading to a misprediction or for a loss of precision due to the over-approximation of the static analysis.
\end{enumerate}

Since our approach relies on sound static analysis engines, it is not possible to miss attacks, i.e., every instance which can be mispredicted upon attack must fall in the second case of our analysis. Let $P^\#(\vec{x}_i) = Y_i$ stand for the set of labels $Y_i$ returned by the analysis of $P$ on the instance $\vec{x}_i$.

By using this information, we can construct an \emph{abstraction} of the behaviour of $t$ under attack on $\dtest$ defined as follows:
\[
\forall (\vec{x}_i,y_i) \in \dtest: t^\#(\vec{x}_i) = 
\begin{cases}
y_i & \text{if } P^\#(\vec{x}_i) = \{y_i\} \\
y \neq y_i & \text{otherwise}
\end{cases}
\]

By construction, we have that $\atkloss(t,\dtest) \leq \Loss(t^\#,\dtest)$ for any loss function which depends just on the number of mispredictions, like the \emph{error rate}, i.e., the fraction of wrong predictions among all the performed predictions. This means that after building $t^\#$ we have an efficient way to over-approximate the loss under attack $\atkloss$ by computing just a traditional loss $\Loss$, which does not require the computation of the set of attacks.


\subsection{Extensions}
\label{sec:extensions}
We discuss here possible extensions of our approach to different popular settings. We leave the implementation of these extensions to future work, since they are essentially an engineering effort.

\subsubsection{Regression.}
The \emph{regression} task requires one to learn a regressor rather than a classifier from the training data. The key difference between a regressor and a classifier is that the former does not assign a class from a finite set $\labels$, but rather infers a numerical quantity from an unbound set, e.g., estimates the salary of an employee based on her features. Regression can be modeled by revising the abstraction $t^\#$ such that it returns an abstract value over-approximating all the values in the predictions found in the leaves which are reachable upon attack. Formally, this means requiring $t^\#(\vec{x}_i) = \sqcup_{y_i \in P^\#(\vec{x}_i)} \alpha(\{y_i\})$, where $\sqcup$ stands for the upper bound operator on the abstract domain.

\subsubsection{Tree Ensembles.}
Ensemble methods train multiple decision trees and combine them to improve prediction accuracy. Traditional ensemble approaches include random forest~\cite{breiman2001randomforest} and gradient boosting~\cite{friedman2001greedy}. Irrespective of how an ensemble is trained, its final predictions are performed just by aggregating the predictions of the individual trees, e.g., using majority voting or averaging. This means that it is possible to readily generalize our analysis technique to ensembles by translating each tree therein and by aggregating their predictions in the generated imperative program.

%% file: implementation.tex
\section{Implementation}
\label{sec:implementation}
\begin{figure}[tb]
	\centering
	\includegraphics[width=\textwidth]{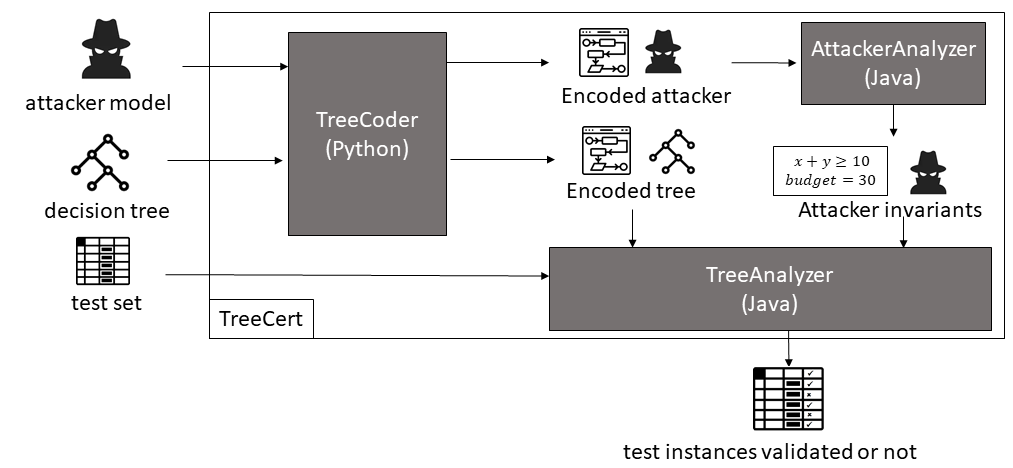}
	\caption{The architecture of TreeCert.}
	\label{fig:architecture}
\end{figure}

Figure~\ref{fig:architecture} depicts the architecture of TreeCert. The inputs are: (i) the attacker, expressed in the threat model of Section~\ref{sec:threat} using a JSON file, (ii) a decision tree to analyse, serialized through the \texttt{joblib} library, and (iii) a test set in CSV format. TreeCert reports for each test instance whether it is always correctly classified for each possible attack or it might be wrongly classified.

\subsection{TreeCoder}
The first step of TreeCert is to encode the attacker and the decision tree as Java programs through the module TreeCoder, as described in Section~\ref{subsect:convertiontoimperativeprogram}. TreeCoder is a Python script that, given a JSON attacker model and a joblib decision tree, produces two distinct Java files encoding the attacker (see method {\sf attack} in Figure~\ref{fig:imp-att}) and the decision tree (see method {\sf predict} in Figure~\ref{fig:imp}). 

There are only two small technical differences over the previous presentation. First, given that all instances of the same dataset share the same set of features, instances are not encoded as arrays, but rather modeled using a distinct local variable for each feature, which simplifies the static analysis; specifically, we let variable ${\sf x}_i$ represent the initial value of the $i$-th feature and variable ${\sf x'}_i$ represent its value after the attack. In addition, each time a rewriting rule $r$ is applied, we increment a counter {\sf r\_counter}, initially set to 0, which allows one to capture useful analysis invariants. Clearly, these changes do not affect the semantics of the generated program, so we did not include them in Figure~\ref{fig:imp-att} for simplicity.


\subsection{AttackerAnalyzer}
The encoded attacker is then passed to the AttackerAnalyzer module, a static analyzer based on abstract interpretation. The analyzer interfaces with Apron, a standard library implementing many popular abstract domains. The analyzer then computes a fixpoint over the Java program representing the attacker, using the Polka implementation\footnote{{\tt http://apron.cri.ensmp.fr/library/0.9.10/mlapronidl/Polka.html}} of the Polyhedra domain~\cite{Polyhedra}.

Polka tracks linear equalities and inequalities over an arbitrary number of variables. These invariants allow AttackerAnalyzer to infer the upper and lower bounds of each attacked feature, based on how many times a feature can be attacked using the available budget. To exemplify, pick the attacker in Figure~\ref{fig:imp-att}. AttackerAnalyzer infers on such program that after the attack has been performed: (i) the value of the first feature may have been decreased by at most $\mathsf{r1\_counter}$ (formally, $\mathsf{x'_1} \in [\mathsf{x_1} - 1 * \mathsf{r1\_counter}, \mathsf{x_1}]$), (ii) the second feature may have been increased by at most $\mathsf{r2\_counter}$ ($\mathsf{x'_2} \in [\mathsf{x_2}, \mathsf{x_2} + 1 * \mathsf{r2\_counter}]$), (iii) both the counters are non-negative ($\mathsf{r1\_counter} \geq 0 \land \mathsf{r2\_counter} \geq 0$), and (iv) the budget spent in the application of the two rewriting rules is less than or equal to the initial budget ($5 * \mathsf{r1\_counter} + 4 * \mathsf{r2\_counter} \leq 10$).
%
%
Note that the last invariant is inferred only if the calculation of a fixpoint over the abstract semantics did not require to apply the Polyhedra widening operator to convergence. Otherwise, the analysis would drop such information to ensure termination.

\subsection{TreeAnalyzer}
The attacker invariants are then passed to the TreeAnalyzer module together with the test set. Like AttackerAnalyzer, TreeAnalyzer performs a static analysis using the Polka implementation of the Polyhedra abstract domain. For each test instance $\vec{x}$, TreeAnalyzer (i) adds to the attacker invariants the initial values of the features of $\vec{x}$, (ii) computes the fixpoint over the program encoding the decision tree $t$ under attack, and (iii) uses it to return the output of $t^\#(\vec{x})$.

To clarify, consider again Example~\ref{ex:example}, where the test instance $(6,8)$ is correctly classified as $-1$ by the decision tree in Figure~\ref{fig:dt}, but can be misclassified upon attack. First of all, TreeAnalyzer adds the invariants $\mathsf{x_1} = 6$ and $\mathsf{x_1} = 8$ to the inferred attacker invariants, leading to an initial Polyhedra state tracking that $\mathsf{x'_1} \in [6-\mathsf{r1\_counter}, 6]$ and $\mathsf{x'_2} \in [8, 8+\mathsf{r2\_counter}]$ with $5 * \mathsf{r1\_counter} + 4 * \mathsf{r2\_counter} \leq 10$. Then the static analysis of the encoded tree starts with the evaluation of the condition $\mathsf{x'_2} \leq 10$, inferring that such condition is always evaluated to true: indeed, $\tt x2$ could be greater than 10 only if {\sf r2\_counter} was strictly greater than 2, but then $5 * \mathsf{r1\_counter} + 4 * \mathsf{r2\_counter} \leq 10$ could not hold since $\mathsf{r1\_counter} \geq 0$. TreeAnalyzer then analyzes the condition $\mathsf{x'_1} \leq 5$. In this case, it cannot definitely conclude that the condition is always evaluated to false, since {\sf x1} can be less than or equal to 5 if $\mathsf{r1\_counter} \geq 1$, which is allowed by the invariant $5 * \mathsf{r1\_counter} + 4 * \mathsf{r2\_counter} \leq 10$. TreeAnalyzer then concludes that the test case might be wrongly classified, since a branch that classifies the test case with +1 could be reached.

%% file: experiments.tex
\section{Experimental Evaluation}
\label{sec:experiments}

\subsection{Methodology}
We evaluate our proposal on three public datasets: Census, House and Wine, which are described in Section~\ref{sec:datasets}.
Our methodology includes multiple steps. We start with a preliminary \emph{threat modeling} phase, where we define the attacker's capabilities by means of a set of rewriting rules $R$ and a set of possible budgets $\{K_1,\ldots,K_n\}$, as explained in Section~\ref{sec:threat}. Our attackers are primarily designed to perform an experimental evaluation of TreeCert, yet they are representative of plausible attack scenarios which do not fit traditional distance-based models and are instead readily supported by the expressiveness of our threat model.

Datasets are divided into $\dtrain$ and $\dtest$ by using 90-10 spitting with stratified sampling (80-20 splitting is used for the smaller Wine dataset).
We first train a decision tree $t$ on $\dtrain$ using the popular \texttt{scikit-learn} library, tuning the maximum number of leaves in the set $\{2^1, 2^2,\ldots, 2^{10}\}$ through cross validation on $\dtrain$. We then evaluate the tree resilience to attacks against each attacker $A = (R,K_i)$ on $\dtest$, using a non-approximated technique. Given the expressiveness of our threat model, the only available solution for this is the algorithm in~\cite{CalzavaraLT19}. In particular, the algorithm computes $\mathbb{A}(\vec{x}_i)$, the set of \emph{representative} attacks against $t$, for each instance $\vec{x}_i$ in $\dtest$. This is a comparatively small subset of the attacks $A(\vec{x}_i)$, which suffices to detect the successful evasions attacks. We refer to this method as {\em Representative Attacks}. We observe and we experimentally confirm that computing even the representative attacks is intractable in general, which motivates the need for approximated analyses, yet being able to do it in a few cases is useful to assess the precision of TreeCert against a ground truth.

Finally, we compute the abstraction $t^\#$ on $\dtest$ for each attacker $A = (R,K_i)$ by using TreeCert. This allows us to classify each $(\vec{x}_i,y_i) \in \dtest$ as follows:
\begin{itemize}
    \item \emph{True Positive} ($\mathit{TP}$): TreeCert states that the instance $\vec{x}_i$ can be misclassified upon attack and this conclusion is correct. Formally, $t^\#(\vec{x}_i) \neq y_i \wedge \exists \vec{x}_i' \in \mathbb{A}(\vec{x}_i): t(\vec{x}_i') \neq y_i$.
    \item \emph{False Positive} ($\mathit{FP}$): TreeCert states that the instance $\vec{x}_i$ can be misclassified upon attack, but this conclusion is wrong. Formally, $t^\#(\vec{x}_i) \neq y_i \wedge \forall \vec{x}_i' \in \mathbb{A}(\vec{x}_i): t(\vec{x}_i') = y_i$.
    \item \emph{True Negative} ($\mathit{TN}$): TreeCert states that the instance $\vec{x}_i$ cannot be misclassified upon attack and this conclusion is correct. Formally, $t^\#(\vec{x}_i) = y_i \wedge \forall \vec{x}_i' \in \mathbb{A}(\vec{x}_i): t(\vec{x}_i') = y_i$.
    \item \emph{False Negative} ($\mathit{FN}$): TreeCert states that the instance $\vec{x}_i$ cannot be misclassified upon attack, but this conclusion is wrong. Formally, $t^\#(\vec{x}_i) = y_i \wedge \exists \vec{x}_i' \in \mathbb{A}(\vec{x}_i): t(\vec{x}_i') \neq y_i$.
\end{itemize}

Since our analysis is sound, we cannot have $\mathit{FN}$. We then assess the quality of TreeCert by computing its \emph{False Positive Rate} $\mathit{FPR}$ and \emph{False Discovery Rate} $\mathit{FDR}$ as follows:
\[
\begin{array}{lll}
\mathit{FPR} = \dfrac{\mathit{FP}}{\mathit{FP} + \mathit{TN}}, & \quad & \mathit{FDR} = \dfrac{\mathit{FP}}{\mathit{FP} + \mathit{TP}}.
\end{array}
\]

We also compare the value of the loss under attack $\atkloss(t,\dtest)$ against its over-approximation $\Loss(t^\#,\dtest)$, focusing on the \emph{error rate}, i.e., the fraction of wrong predictions. 
Finally, we compare the execution times of TreeCert against the time spent in the computation of the set of the representative attacks. 

\subsection{Datasets}
\label{sec:datasets}
We perform our experiments on three publicly available datasets. The preconditions of the rewriting rules and the magnitude of the perturbations have been set after a preliminary data exploration step, based on the observed data distribution in the dataset. Statistics about the datasets are in Table~\ref{tab:datasets}.

\subsubsection{Census.}
The Census\footnote{\url{http://archive.ics.uci.edu/ml/machine-learning-databases/adult/adult.data}} dataset includes demographic information about American citizens. The prediction task is estimating whether the income of a citizen is above 50,000\$ per year. For this dataset, we define four rewriting rules: 
\begin{itemize}
\item cost 5: if the capital gain is in [0,100000], a citizen can raise it by 200;
\item cost 5: if the capital loss is in [0,100000], a citizen can lower it by 200;
\item cost 10: if the number of work hours is in [0,40], a citizen can raise it by 1;
\item cost 10: if the age is in [0,40], a citizen can raise it by 1. 
\end{itemize}
We consider 20, 40, 60, 80 as possible values of the attacker's budget.

\begin{table}[t]
     \centering
\caption{\label{tab:datasets} Properties of datasets used in the experiments.}
\bgroup
\renewcommand{\tabcolsep}{.5em}
\renewcommand{\arraystretch}{1.1}
\begin{tabular}{c|ccc}
Dataset 	& \#Instances 	& \#Features 	& Maj. class 	\\
\hline
Census 		& 29169 	&  51 		& 0.75 	\\
House 		& 21613 	&  19 		& 0.51 	\\
Wine 		&  6497 	&  12 		& 0.63 	\\
\hline
\end{tabular}
\egroup
\end{table}

\subsubsection{House.}
The House\footnote{\url{https://www.kaggle.com/harlfoxem/housesalesprediction}} dataset contains house sale prices for the King County area. The prediction task is inferring whether a house costs at least the median house price. For this dataset, we define four rewriting rules:
\begin{itemize}
    \item cost 5: if the square footage of the living space of the house is in [0,3000], it can be increased by 50;
    \item cost 5: if the square footage of the land space is in [0,2000], it can be increased by 50;
    \item cost 5: if the average square footage of the living space of the 15 closest houses is in [0,2000], it can be increased by 50;
    \item cost 5: if the construction year is in [1900,1970], it can be increased by 10.
\end{itemize}
We consider 10, 20, 30, 40 as possible values of the attacker's budget.

\subsubsection{Wine.}
The Wine\footnote{\url{https://www.openml.org/data/get\_csv/49817/wine\_quality.arff}} dataset represents different types of wines. The prediction task is detecting whether a wine has quality score at least 6 on a scale 0--10. For this dataset, we define four rewriting rules:
\begin{itemize}
\item cost 2: if the residual sugar is in [2,4], it can be lowered by 0.01;
\item cost 5: if the alcohol level is in [0,11], it can be increased by 0.01;
\item cost 5: if the volatile acidity is in [0,1], it can be lowered by 0.01;
\item cost 5: if the free sulfur dioxide is in [20,40], it can be lowered by 0.1.
\end{itemize}
We consider 20, 30, 40, 50, 60 as possible values of the attacker's budget.

\subsection{Experimental Results}

\subsubsection{Precision.} Table~\ref{tab:results} reports for all datasets and budgets a number of measures computed for the trained decision tree $t$:
\begin{enumerate}
    \item the traditional loss in absence of attacks $\Loss(t,\dtest)$. This is the fraction of wrong predictions returned by $t$ on $\dtest$ in the unattacked setting;
    \item the loss under attack $\atkloss(t,\dtest)$, computed by enumerating all the representative attacks using the algorithm in~\cite{CalzavaraLT19}. This is the fraction of wrong predictions returned by $t$ on $\dtest$ upon attack;
    \item the over-approximation of the loss under attack $\Loss(t^\#,\dtest)$, computed using the program analysis of TreeCert;
    \item the false positive rate of TreeCert, noted $\mathit{FPR}$;
    \item the false discovery rate of TreeCert, noted $\mathit{FDR}$.
\end{enumerate}

\begin{table}[t]
     \centering
\caption{\label{tab:results} Accuracy results across dataset.}
\bgroup
\renewcommand{\tabcolsep}{.5em}
\renewcommand{\arraystretch}{1.1}

\begin{tabular}{c|c|ccc|cc}
Dataset   & Budget & $\Loss(t,\dtest)$   & $\atkloss(t,\dtest)$   & $\Loss(t^\#,\dtest)$ & $\mathit{FPR}$    & $\mathit{FDR}$  	\\
\hline\hline
\multirow{ 4 }{*}{ Census }
 	& 20 	& 0.14 	& 0.17 	& 0.17 	& 0.00 	& 0.00 	\\
 	& 40 	& 0.14 	& 0.17 	& 0.17 	& 0.00 	& 0.01 	\\
 	& 60 	& 0.14 	& 0.18 	& 0.18 	& 0.00 	& 0.01 	\\
 	& 80 	& 0.14 	& 0.20 	& 0.21 	& 0.00 	& 0.01 	\\
\hline\hline
\multirow{ 4 }{*}{ House }
 	& 10 	& 0.10 	& 0.12 	& 0.12 	& 0.00 	& 0.02 	\\
 	& 20 	& 0.10 	& 0.14 	& 0.15 	& 0.01 	& 0.04 	\\
 	& 30 	& 0.10 	& 0.16 	& 0.17 	& 0.01 	& 0.06 	\\
 	& 40 	& 0.10 	& 0.18 	& 0.19 	& 0.02 	& 0.08 	\\
\hline\hline
\multirow{ 5 }{*}{ Wine }
 	& 20 	& 0.24 	& 0.30 	& 0.31 	& 0.01 	& 0.02 	\\
 	& 30 	& 0.24 	& 0.34 	& 0.35 	& 0.02 	& 0.03 	\\
 	& 40 	& 0.24 	& 0.36 	& 0.37 	& 0.02 	& 0.04 	\\
 	& 50 	& 0.24 	& 0.37 	& 0.39 	& 0.03 	& 0.05 	\\
 	& 60 	& 0.24 	& 0.38 	& 0.40 	& 0.03 	& 0.05 	\\
\hline
\end{tabular}

\egroup
\end{table}

The experimental results clearly confirm the quality of the analysis performed by TreeCert. In particular, we observe that the $\mathit{FPR}$ is remarkably low, standing well below 5\%, where 10\% is considered a state-of-the-art reference for static analysis techniques~\cite{Google}. 
Indeed, in Census we measured an absolute number of false positives never greater than 5.
This is interesting, because it shows that for many instances there is a simple security proof, i.e., TreeCert is able to prove that they cannot be successfully attacked (i.e., they are $\mathit{TN}$), which significantly drops the $\mathit{FPR}$. As to the $\mathit{FDR}$, we observe that it also scores extremely well on all datasets, though it tends to be slightly higher than $\mathit{FPR}$. However, this is not a major problem in our application setting: contrary to what happens in traditional program analysis, where users are forced to investigate all false alarms to identify possible bugs, here we are rather interested in the aggregated analysis results, i.e., the final over-approximation of the loss under attack. Even on the House dataset, where $\mathit{FDR}$ tends to be higher, we observe that the loss under attack is appropriately approximated by TreeCert, since there is a difference of at most 0.01 between the actual value and its over-approximation. Remarkably, our experiments also show that the quality of the over-approximation is not significantly affected by the attacker's budget, which is important because it suggests that TreeCert likely generalizes to cases where computing the actual value of the loss under attack is computationally intractable, which is the intended use case of such an analysis tool.

\subsubsection{Efficiency.} 
\begin{figure}[t]
    \centering
    \includegraphics[width=\textwidth]{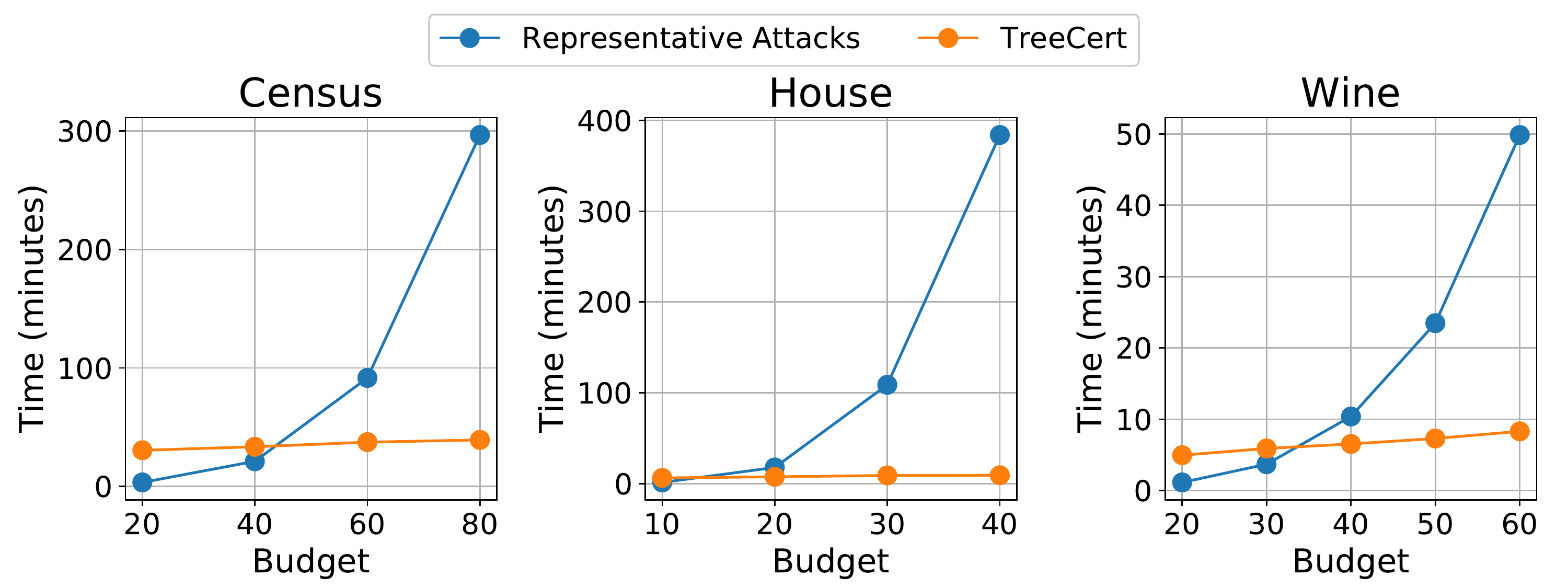}
    \caption{Running time of TreeCert against the enumeration of representative attacks.}
    \label{fig:time}
\end{figure}

To show the efficiency of our approach, we compare in Figure~\ref{fig:time} the running time of TreeCert against the time taken to compute the full set of the representative attacks. It is possible to clearly see that the two curves exhibit completely different trends. The time taken to construct the representative attacks has an \emph{exponential} trend: the approach is efficient and feasible when the attacker's budget is low, but blows up to intractability very quickly. For example, each increase in the attacker's budget multiplies the execution time of a 3x factor in the case of Census and we experimentally confirmed that more than 12 hours of computation are needed when the budget grows to 100 (not plotted). Conversely, the execution time of TreeCert is only marginally affected when increasing the attacker's budget, since the analysis always converges in less than one hour. In the case of the House dataset, computing the set of the representative attacks is even less feasible: even for small budgets, the running time is remarkably high, due to the fact that the trained decision tree uses many different thresholds, which makes the number of representative attacks blow up. Finally, also the Wine dataset shows similar figures, though the execution times there are lower due to its smaller size. This confirms that brute-force approaches based on the exhaustive enumeration of the representative attacks do not scale, yet luckily they can be replaced by more efficient abstraction techniques with good precision.

%% file: related.tex
\section{Related Work}
\label{sec:related}
Verifying the security guarantees of machine learning models is an important task, which received significant attention by the research community in the last few years. In particular, many papers proposed techniques to verify the security of deep neural networks~\cite{WangPWYJ18,WangPWYJ18b,HuangKWW17,KatzBDJK17,GehrMDTCV18}; we refer to a recent survey for more work in this research area~\cite{abs-1810-01989}. As of now, however, comparatively less attention has been received by the security verification of decision trees models.

The closest related work to our approach is a very recent paper by Ranzato and Zanella~\cite{RanzatoZ20}. Their work also focuses on decision trees and builds on the abstract interpretation framework. However, their approach can only be applied to attackers who admit a simple mathematical characterization as a set of perturbations, e.g., based on distances. In particular, their soundness theorem relies on the hypothesis that, for each test instance $\vec{x}$, one has $A(\vec{x}) \subseteq \gamma(\alpha(\{\vec{x}\}))$, i.e., the abstraction of $\vec{x}$ must cover all the possible attacks. Checking this condition for distance-based attackers is straightforward, yet it is computationally infeasible in general. For example, in the case of the rewriting rules we considered, $A(\vec{x})$ is unknown \emph{a priori}, but is induced by the application of the rules. Indeed, their tool \emph{silva} only supports attackers based on the infinity-norm $L_\infty$, which has a compact mathematical characterization as a set, but falls short of representing realistic threats. Instead, our approach is general enough to work on attackers modeled as arbitrary imperative programs.

Other approaches also deal with the verification of decision trees, but are not based on abstract interpretation. For example, Einzinger et al. use SMT solving to verify the robustness of gradient-boosted models~\cite{EinzigerGSS19}. Their approach also requires to explicitly encode the set of attacks $A(\vec{x})$ in closed form, which is only easily doable for artificial distance-based attackers. Moreover, SMT solving suffers from scalability issues, which required the authors to develop custom optimizations to make their approach practical. It is unclear whether this line of work can be adapted and scale to more expressive attackers or not, also because their tool is not publicly available. Other notable work include the robustness verification algorithm by Chen et al.~\cite{ChenZS0BH19}, which only works for attackers based on the infinity-norm $L_\infty$, and the abstraction-refinement approach by Törnblom and  Nadjm-Tehrani~\cite{TornblomN19}, which is not proved sound.

Finally, it is worth mentioning adversarial learning algorithms which train decision trees more resilient to evasion attacks by construction~\cite{KantchelianTJ16,CalzavaraLT19,ChenZBH19,CalzavaraLTAO19}. This line of work is orthogonal to the security verification of decision trees, i.e., our approach can also be applied to estimate the improved robustness guarantees of trees trained using such algorithms.

%% file: conclusion.tex
\section{Conclusion}
We proposed a technique to certify the security of decision trees against evasion attacks by leveraging the abstract interpretation framework. This is the first solution which is both sound and expressive enough to deal with sophisticated attackers represented as arbitrary imperative programs. Our experiments showed that our technique is both precise and efficient, yielding only a minimal number of false positives and scaling up to cases which are intractable for a competitor~\cite{CalzavaraLT19}.

We foresee several avenues for future work. First, we plan to extend our approach to the analysis of regression tasks and tree ensembles: though this is straightforward from an engineering perspective, we want to analyze the precision and the efficiency of our solution in such settings. Moreover, we will investigate techniques to automatically infer the minimal attacker's budget required to induce a given error rate on the test set, so as to efficiently provide security analysts with this useful information. Finally, we will investigate the trade-off between the precision and the efficiency of TreeCert by testing more sophisticated abstract domains and analysis techniques, e.g., trace partitioning.